\documentclass[a4paper,fleqn]{cas-dc}

\usepackage[authoryear,longnamesfirst]{natbib}
\usepackage{algorithm}

\usepackage{graphicx,subfigure}

\def\tsc#1{\csdef{#1}{\textsc{\lowercase{#1}}\xspace}}
\tsc{WGM}
\tsc{QE}

\begin{document}
\let\WriteBookmarks\relax
\def\floatpagepagefraction{1}
\def\textpagefraction{.001}

\shorttitle{Low Precision Decentralized Training}    

\shortauthors{S.A. Aketi et al.}  

\title [mode = title]{Low Precision Decentralized Distributed Training over IID and non-IID Data}  



%

\author[1]{Sai Aparna Aketi}[orcid=0000-0003-3446-0243]

\cormark[2]


\ead{saketi@purdue.edu}



\affiliation[1]{organization={Purdue University},
            addressline={}, 
            city={West Lafayette},
            postcode={47906}, 
            state={IN},
            country={USA}}
\author[1]{Sangamesh Kodge}[]


\ead{skodge@purdue.edu}


\author[1]{Kaushik Roy}[]


\ead{kaushik@purdue.edu}



\cortext[2]{Sai Aparna Aketi}



\begin{abstract}
Decentralized distributed learning is the key to enabling large-scale machine learning (training) on the edge devices utilizing private user-generated local data, without relying on the cloud. However, practical realization of such on-device training is limited by the communication and compute bottleneck. In this paper, we propose and show the convergence of low precision decentralized training that aims to reduce the computational complexity and communication cost of decentralized training. Many feedback-based compression techniques have been proposed in the literature to reduce communication costs. To the best of our knowledge, there is no work that applies and shows compute efficient training techniques such as quantization, pruning etc., for peer-to-peer decentralized learning setups. Since real-world applications have a significant skew in the data distribution, we design "Range-EvoNorm" as the normalization activation layer which is better suited for low precision training over non-IID data. Moreover, we show that the proposed low precision training can be used in synergy with other communication compression methods decreasing the communication cost further. Our experiments indicate that 8-bit decentralized training has minimal accuracy loss compared to its full precision counterpart even with non-IID data. However, when low precision training is accompanied by communication compression through sparsification we observe a $1-2\%$ drop in accuracy. The proposed low precision decentralized training decreases computational complexity, memory usage, and communication cost by $\sim 4\times$ and compute energy by a factor of $\sim 20 \times$, while trading off less than a $1\%$ accuracy for both IID and non-IID data. In particular, for higher skew values, we observe an increase in accuracy (by $\sim 0.5\%$) with low precision training, indicating the regularization effect of the quantization. 
\end{abstract}




\maketitle

\section{Introduction}
\label{intro}

Deep learning models have achieved exceptional performance in many computer vision, natural language processing (NLP) and reinforcement learning (RL) tasks \cite{cv, nlp, rl}. This remarkable success is largely attributed to the availability of humongous amounts of data and computational power \cite{vit,sanh2019distilbert}. These models are traditionally trained on a single system or a cluster of nodes by centralizing data from various distributed sources or edge devices. This results in transferring tremendous amount of information to data-centers or the cloud and subsequently train the models using enormous computational power. Such centralized computing, even though successful in some industrial use cases, requires large amount power and network bandwidth for transfer of data from edge devices \cite{li2017multi} and also increases users' privacy concerns \cite{swinhoe202015}. 
This fundamentally limits the amount of data that can be collected and used for training the more intelligent and generalized models. In order to handle this problem, a new interest in developing distributed learning algorithms has emerged. 
A well-known distributed optimization setting in machine learning is federated learning or centralized distributed learning \cite{konevcny2016federated}. The idea here is to keep the training data locally at the edge devices and learn shared global model by aggregating the locally computed updates through a central coordinating server. This avoids the logging of user data to a data center and helps safeguard users' privacy. However, communication traffic jam at the central server and a single point of failure becomes a potential bottleneck for such centralized settings. This has motivated the advancements in decentralized distributed learning algorithms where each node communicates only with its neighbours eliminating the requirement of a central server \cite{dpsgd}.

\textit{Key Challenges in On-Device Decentralized Learning:} There are three main challenges in realizing decentralized training on edge devices. Training a model over decentralized peer-to-peer learning using traditional algorithms such as D-PSGD \cite{dpsgd} requires massive amounts of communication. Second, deep learning models are usually very compute heavy and training such models on resource constrained edge devices is infeasible. Lastly, the training data on each device is based on the behaviour and preference of the user and this can lead to significant differences in the data distribution across the devices depending on the user pool. However, most of the existing works \cite{deepsqueeze, choco2, chocosgd, moniqua, quant_sgp} only focus on communication bottleneck and analyze them assuming the data partitions to be independent and identically distributed (IID). 
This leaves us with an important question: \textit{What is the effect of compute efficient low precision training on communication efficient peer-to-peer decentralized machine learning systems?}

In this paper, we aim to show that low precision (8-bit) training methodology proposed for single node \cite{8bit} will converge for decentralized systems with some modifications to the normalization-activation layers. 
Since `Batch-Norm' layer is not well suited for low precision training \cite{8bit} and for decentralized learning with non-IID data \cite{skewscout}, we propose `Range EvoNorm' by combining the concepts of `Range Batch-Norm' \cite{8bit} and S0 variant of `EvoNorm' \cite{evonorm}.
We evaluate the proposed low precision training on two popular communication efficient decentralized algorithms, Deep-Squeeze \cite{deepsqueeze} and Choco-SGD \cite{choco2}, over various applications and architectures. Since the above mentioned algorithms converge only for undirected graphs, we modify them to include Push-Sum technique \cite{sgp} similar to \cite{sparsepush, quant_sgp} removing the constraint of requiring symmetric and doubly stochastic graph topologies. Further, we show that the low precision training can be used in synergy with communication compression techniques such as layer-wise top-k sparsification.
Note that \textit{Layer-wise top-K sparsification} indicates that only top $K\%$ (in terms of magnitude) of the weights/gradients of each layer are communicated to the neighbouring nodes.
We demonstrate the empirical convergence of the proposed low precision decentralized training over both IID and non-IID data.
We mainly focus on a common type of non-IID data, widely used in prior work \cite{skewscout, zhao2018federated, pmlr-v80-tang18a}: skewed distribution of data labels across devices. 

Our experiments show that the low precision decentralized training has minimal drop in accuracy as compared to its full precision counter part. The degree of non-IIDness (skew) in the data partitions has very similar effect on low-precision and full-precision training. 
We further observe that, under high degree of skew, the proposed low precision decentralized training with optimal models (e.g., ResNet-20 for CIFAR-10) improves accuracy slightly by $\sim 1\%$ compared to full precision training. We hypothesise that the improvement is due to the regularization effect of the low precision training.
We further replace the local momentum with Quasi-Global (QG) momentum as suggested in \cite{qgm} to improve the accuracy by $1-2\%$ for decentralized setups with non-IID data. 

\subsection{Contributions:}
We make the following contributions.
\begin{itemize}
    \item We propose `Range EvoNorm' which is better suited non-linearity for low precision decentralized training over both IID and non-IID data.
    It replaces the variance (of activations computer per instance) term in the S0 variant of EvoNorm \cite{evonorm} with a scaled range of the activations, and sigmoid non-linearity by hard-sigmoid activation.
    \item We study the feasibility of low precision decentralized training with the proposed Range EvoNorm over both IID and non-IID data distributions on various image classification and natural language processing tasks (refer Section \ref{sec:exps}).  To the best of our knowledge, this is the first time that low precision training has been applied to decentralized learning setup.
    \item We show that the low precision training can be used in synergy with the communication compression techniques such as top-k sparsification with error feedback. This reduces the communication cost further, adding to the $4\times$ reduction due to quantized weights Table \ref{tab:cifar}. Moreover, the proposed technique can reduce the energy requirement by $20 \times$ (refer Section \ref{sec:hardware}). 
\end{itemize}

\section{Background}
\label{bg}

In this section, we provide the background on decentralized peer-to-peer learning algorithms.

\textit{Decentralized Learning:} In decentralized learning, the aim is to train a machine learning model with parameter vector $x$ using all the training samples that are generated and stored across $n$ edge devices. The end goal is to learn one model that fits all the training samples without sharing the local data. This is usually achieved by combining a local stochastic gradient descent with global consensus based optimization such as gossip averaging \cite{dpsgd}. The nodes/devices are connected via a weighted, sparse, yet strongly connected graph topology.
Traditionally, most decentralized learning algorithms assume the data samples to be independent and identically distributed (IID) among different devices, and this is referred as the IID setting. Conversely, we refer non-IID settings as the one for which the above assumption does not hold.

In this paper, we show the convergence of decentralized low precision training by evaluating it using two popular algorithms on different applications over IID and Non-IID settings. Both the algorithms use gossip based global averaging with stochastic gradient descent and aim to reduce communication by sending compressed model weights or gradients to the neighbours. The following are the details of these algorithms.
\begin{itemize}
    \item \textit{Deep-Squeeze} \cite{deepsqueeze} employs error compensated communication compression to a gossip based decentralized learning setup. Without communication compression, Deep-Squeeze is the same as Decentralized Parallel Stochastic Gradient Descent (D-PSGD) proposed in \cite{dpsgd}. Any given device (say $i$) updates its model weights $x_i$ locally using SGD and then communicates only the top-$k\%$ of the error compensated weights to the neighbours. Every local update step is followed by a gossip averaging step where each node updates the model with a weighted average of all the communicated models (from its neighbours) including itself. 
    (Algorithm 2 in the Supplementary material)
   
    \item \textit{Choco-SGD} \cite{choco2} is another gossip-based stochastic gradient descent algorithm that communicates compressed model updates (i.e. $x_i^{(t)}-x_i^{(t-1)}$) rather than the model weights. They achieve this by introducing an additional buffer $\hat{x_i}$ that acts as a proxy to the model weights $x_i$. The $\hat{x_i}$'s are available to all the neighbours of the node $i$ and referred as the ‘publicly available’ copies of private $x_i$. The compressed gradients received from a neighbor $j$ to $i$ are used to update $\hat{x_j}$ at node $i$ and the gossip averaging step averages $x_i$ and $\hat{x_j}$'s. In general $x_i \neq \hat{x_i}$, due to the communication restriction. 
    Since Choco-SGD communicates only gradients (in particular, model updates) and the gradients converge to zero as the training converges, we expect it to performs better than Deep-squeeze.  
    (Algorithm 3 in the Supplementary material)
\end{itemize}

Both the above mentioned algorithms assume the data partitions to be IID and the mixing matrix (adjacency matrix $W$) of the decentralized setup to be symmetric and doubly stochastic. Doubly stochastic constraint requires every row and column to sum to $1$.
The authors in \cite{sparsepush} propose Sparse-Push combining deep-squeeze algorithm with Stochastic Gradient Push (SGP) \cite{sgp} which allows the training to converge for directed and time varying graphs reducing the constraint on mixing matrix to be only column stochastic. Quant-SGP \cite{quant_sgp} extended D-PSGD to directed graphs by effectively combining SGP algorithm with Choco-SGD. In this paper, we will show the convergence of low precision decentralized training over directed graphs with non-IID data using Quant-SGP variant of Choco-SGD and Sparse-Push variant of Deep-Squeeze.

\section{Decentralized Low Precision Training}
In this section, we describe our low precision training methodology for decentralized learning setup over IID and non-IID data partitions.

\textit{Decentralized Setup:} We consider a decentralized peer-to-peer learning problem with $n$ nodes connected over a directed and weighted graph topology $G=(n,E)$ where $E$ are the set of edges. The edge $(i,j) \in E$ indicates that node $i$ communicates information to node $j$. We represent $N(i)$ as the in-neighbours of $i$ including itself i.e. set of nodes with incoming edge to $i$. Additionally, we assume that the graph $G$ is strongly-connected (there is a path from every node to every other node) and has self-loops. The adjacency matrix of the graph $G$ is referred as a mixing matrix $W$ where $W_{ij}$ is the weight associated with edge $(i,j)$. The values of weights lie in the interval of $[0,1]$. Note that a weight 0 indicates the absence of an edge and a non-zero weight indicate the weightage given to model at node $j$ while being averaged at node $i$.
The weight matrix is assumed to be column stochastic i.e. the columns sum to $1$. Further, each node has its own local data coming from distribution $D_i$ and an initial model estimate $x_i^{(0)}$. 
Note that even though the learning set-up is peer-to-peer, we assume (similar to all previous works in the literature) that the initial models and all hyperparameters are synchronized in the beginning of the training. 
Our aim is to solve the optimization problem of minimizing global loss function $f(x)$ distributed across $n$ nodes as given in equation.~\ref{eq:1}. Note that $F_i$ is a local loss function at node $i$ (for example, cross entropy loss). 
\begin{equation}
\label{eq:1}
\begin{split}
    \min \limits_{x \in \mathbb{R}^d} f(x) &= \frac{1}{n}\sum_{i=1}^n f_i(x), \\
    and \hspace{2mm} f_i(x) &= \mathbb{E}_{\xi_i \in D_i}[F_i(x_i, \xi_i)] \hspace{2mm} \forall i 
\end{split}
\end{equation}

\textit{Decentralized Training:} Communication efficient decentralized training algorithms such as Choco-SGD and Deep-Squeeze usually have four stages in each iteration of the training as shown in Algorithm.~\ref{alg1}. (1) \emph{Local computation stage:}  Compute the stochastic gradient descent updates for each node using a randomly selected batch from the local data. 
(2) \emph{Compression stage:} Compress the model weights/model-updates based on a given compression operator. In case of Deep-Squeeze, error compensated weights (models weights with accumulated compression error) are compressed. In contrast, Choco-SGD compresses the change in model weights from the previous communication step.
(3) \emph{Communication stage:} The compressed information from each node $i$ is sent to its neighbors, namely $N(i)$.
(4) \emph{Gossip update stage:} Each node updates its local model by performing a gossip update utilizing the information received from its neighbors -- a neighbourhood weighted averaging.
The predefined function $g,h$ introduced in Algorithm.~\ref{alg1} are described in detail for Deep-Squeeze and Choco-SGD in supplementary material.

\begin{algorithm}[ht]
\textbf{Input:} Each node $i$ initializes model weights $x_i^{(0)}$, learning rate $\gamma$, averaging rate $\eta$, mixing matrix $W$, and compression operator $C$, predefined functions $g,h$.\\
1. \textbf{for} t=$0,1,\hdots,T-1$ \textbf{do}\\
 \hspace*{3mm} Each node simultaneously implements:\\
2.\hspace*{4mm} Compute the local gradients: $g_i^{(t)}=\nabla F_i(x_i^{(t)}, \xi_i^{(t)}) $\\
3.\hspace*{4mm} Update the model: $\widetilde{x}_i^{(t)}=x_i^{(t)}-\eta g_i^{(t)}$\\
4.\hspace*{4mm} Compute $v_i^{(t)} = g(\widetilde{x}_i^{(t)})$ that has to be communicated.\\
5.\hspace*{4mm} Compress the information $C[v_i^{(t)}]$\\
6.\hspace*{4mm} Communicate $C[v_i^{(t)}]$ to neighbors $N(i)$.\\
7.\hspace*{4mm} Update the model by weighted gossip averaging step:\\
\hspace*{8mm} $x_i^{(t+1)}=\widetilde{x}_i^{(t)}+\gamma\sum_{j\in \mathcal{N}(i)}W_{ij}*h(C[v_{j}^{(t)}])$\\
8. \textbf{end}
\caption{Decentralized Peer-to-Peer Training}
\label{alg1}
\end{algorithm}

\textit{Low Precision Training:} We extend the single node 8-bit training proposed in \cite{8bit} to peer-to-peer decentralized learning setups. This training methodology uses GEMMLOWP integer quantization scheme as described in Google’s open source library \cite{gem}.
In every iteration of training, we quantize the weights, activations, as well as a substantial volume of the gradients stream, in all layers including normalization layers to 8-bit across all nodes. The back-propagation phase requires twice the number of multiplications compared to the forward pass and hence, quantization of gradients is crucial for accelerating the training. Quantizing the entire backward phase results in significant degradation of accuracy and it has been empirically shown in \cite{8bit} that gradient bifurcation during 8-bit training is critical for high accuracy results. Each layer derives two sets of gradients for the weight update i.e., the layer gradients ($g_l$) for the backward propagation that are passed to the previous layer and the weight gradients ($g_{W_l}$) that are used to update the weights in the current layer. In gradient bifurcation, only the layer gradients $g_l$ are quantized to 8-bits integers while the weight gradients $g_{W_l}$ are either 16-bit or 32-bit float. In our implementation, gradient bifurcation takes place during the backward pass at each node. Similar to 8-bit single node training in \cite{8bit}, we used the most simple and hardware friendly approach of straight-through estimator (STE) to approximate differentiation through discrete variables.

Other computations, apart from gradients that can cause numerical instability, are the batch-normalization layers. 
\begin{equation}
\label{eq:bn}
\begin{split}
    BatchNorm: &BN(x)=\frac{x-\mu_{b,w,h}(x)}{\sqrt{s^2_{b,w,h}(x)}}\gamma+\beta\\
    BN\_RELU(x) &=\max\left( \frac{x-\mu_{b,w,h}(x)}{\sqrt{s^2_{b,w,h}(x)}}\gamma+\beta, 0\right)\\
\end{split}
\end{equation}
A traditional implementation of batch-norm (shown as BN in equation.~\ref{eq:bn}) includes the computation of the sum of squares, square-root and reciprocal operations which require high precision and high dynamic range. 
Note that, for all the below equations, $\mu_{b,w,h}$, $s^2_{b,w,h}$ indicates that mean and variance are computed per channel across all the elements $w,h$ and all the samples in a batch of size $b$.
A replacement for batch-norm layer is range batch-norm (shown as $Range\_BN$ in equation.~\ref{eq:rbn}), that normalizes inputs by the range of the input distribution, i.e., $range(v) = \max(v)- \min(v)$, across the batch rather than using the batch variance. 
\begin{equation}
\label{eq:rbn}
\begin{split}
    Range\_BN(x) &= \frac{x-\mu_{b,w,h}(x)}{C(N)*range(x-\mu_{b,w,h}(x))}\gamma+\beta\\
    C(b) &= \frac{1}{\sqrt{2*ln(b)}}, \hspace{2mm} \text{ b = batch size}
\end{split}
\end{equation}
The range version of batch-norm  has been shown to be more suitable for low precision implementations in \cite{8bit} compared to it's traditional implementation. 

\begin{table}[]
    \centering
        \caption{Description for the notations used in the equations.~\ref{eq:bn}, \ref{eq:rbn}, \ref{eq:en}, \ref{eq:ren}}
    \label{tab:notations}
         
    \begin{minipage}{\linewidth}
     \centering
    \resizebox{\textwidth}{!}{
     \begin{tabular}{|c|c|}
         \hline
        Notation  & Description \\
        \hline
        $\mu(x)$  & mean \cite{} \\
        \hline
        $s^2(x)$  & variance \cite{} \\
        \hline
        $range(x)$  & max(x)-min(x)\\
        \hline
        $\sigma(x)$  & sigmoid \cite{}\\
        \hline
        $\hat{\sigma}(x)$  & hard sigmoid \cite{}\\
        \hline
         $\gamma$, $\beta$   & trainable parameters\\
         & of affine transform\\
        \hline
        $w$  & width of the input \\
        \hline
        $h$  & height of the input \\
        \hline
        $c$  & number of input channels \\
        \hline
        $g$  & group size (default=2) \\
        \hline
        $b$  & batch size \\
        \hline
         \end{tabular}
        }
    \end{minipage} 
\end{table}

However, in decentralized learning setup with non-IID data, we observe that the batch-norm often fails due to the discrepancies between local activation statistics across the nodes. This has also been demonstrated in several works such as \cite{skewscout, qgm}. As a counteractive measure, the authors in \cite{qgm} propose to replace batch-norm and ReLU layers (shown as $BN\_ReLU$ in equation.~\ref{eq:bn}) with S0 variant of EvoNorm (shown as EvoNorm in equation.~\ref{eq:en}) proposed by \cite{evonorm}. 
\begin{equation}
\label{eq:en}
\begin{split}
    EvoNorm(x) &= \frac{x\sigma(vx)}{\sqrt{s^2_{w,h,c/g}(x)}}\gamma+\beta\\
    \sigma(x) &= \frac{1}{1+e^{-x}}\\
\end{split}
\end{equation}
As EvoNorm uses sample statistics rather than run-time batch statistics, it shows significant improvement in decentralized settings with non-IID data. EvoNorm divides the channels of the input into groups based on the predefined group size and the variance is computed for each group of channels per sample. 
Note that, for all the below equations, $s^2_{w,h,c/g}$ indicates that variance is computed per sample across all $w,h$ elements in each $c/g$ channels.

\begin{figure}[t]
  \centering
   \includegraphics[width=\linewidth]{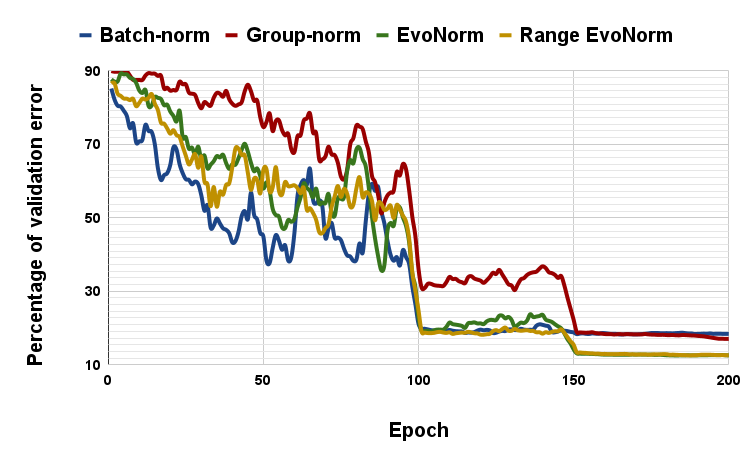}
   \caption{Plot showing the validation error during training of CIFAR-10 dataset on ResNet-20 architecture with various normalization layers over 8 node directed ring topology and a degree of skew of 0.8.}
   \label{fig:norm}
   \vspace{1mm}
\end{figure}

Combining the concepts of range batch-norm and Evo-Norm, we propose `Range EvoNorm' that normalizes inputs by the range of activation distribution of each input. The proposed normalization-activation layer is shown in equation.~\ref{eq:ren}. 
The `Range EvoNorm' replaces the variance computed across group of channels for each sample in the batch with the range. We use similar scaling constant as range batch-norm but it depends on number of channels per group rather than the batch-size.
We also replace the sigmoid function with hard-sigmoid to avoid the computation of exponential function which requires high precision. 
\begin{equation}
\label{eq:ren}
\begin{split}
    Range\_Evo&Norm(x) = \frac{x\hat{\sigma}(vx)}{C(c/g)*range(x_{w,h,c/g})}\gamma+\beta\\
    C(c/g)=&\frac{1}{\sqrt{2*ln(c/g)}}, \hspace{2mm} \text{ c/g = \#channels per group}\\
    \hat{\sigma}(x)&= \max\left(0,\min\left(1,\frac{x+3}{6}\right)\right)
\end{split}
\end{equation}
This ensures that the normalization layer uses sample statistics rather than batch statistics during training. It also avoids computing sum of squares and square-roots which need high precision operations. 
Figure.~\ref{fig:norm} shows the equivalence of `EvoNorm' and `Range EvoNorm' when models are trained on high precision on a decentralized setup. It also demonstrates that `EvoNorm' perform better than batch-norm and group-norm for decentralized settings over non-IID data.
Hence all the models in our experiments shown in the following section use `Range EvoNorm' as the normalization activation layer for low precision training and `EvoNorm' for full precision training. 

\section{Experimental Setup and Results} \label{sec:exps}
We study three different dimensions to evaluate the decentralized low precision training: 1) Communication efficient decentralized learning algorithms such as CHOCO-SGD \cite{chocosgd} and Deep-Squeeze \cite{deepsqueeze} 2) Degree of skew: the amount of non-IIDness present in the data distributions across the nodes and 3) various ML applications and models. We explore all the dimensions with rigorous experimental methodologies. We experiment on two dominant ML applications -- image recognition and NLP. For all the experiments we replace batch-norm + ReLU layers with $S0$ variant on EvoNorm \cite{evonorm} for full precision training and Range EvoNorm for low precision 8-bit training. We used layer-wise top-K sparsification compressor for decentralized training experiments with communication compression.

The baseline accuracies for all the datasets and architectures trained on a single node with entire dataset are reported in Table.~\ref{tab:baseline}. In the tables \ref{tab:cifar}, \ref{tab:qgm} \ref{tab:imagenette}, \ref{tab:nlp}: DS, LP-DS indicates Deep-Squeeze (full precision) \cite{deepsqueeze} and our Low Precision Deep-Squeeze respectively; Choco and LP-Choco represents Choco-SGD (full precision) \cite{choco2} and our Low Precision Choco-SGD, respectively; QGM is Quasi-Global Momentum \cite{qgm}. The performance is measured in terms of the test accuracy of the averaged model across all nodes. 
\footnote{Our PyTorch code is available at \url{https://github.com/aparna-aketi/Low_Precision_DL}} 

\subsection{Computer Vision Tasks}
\begin{figure}[t]
  \centering
   \includegraphics[width=\linewidth]{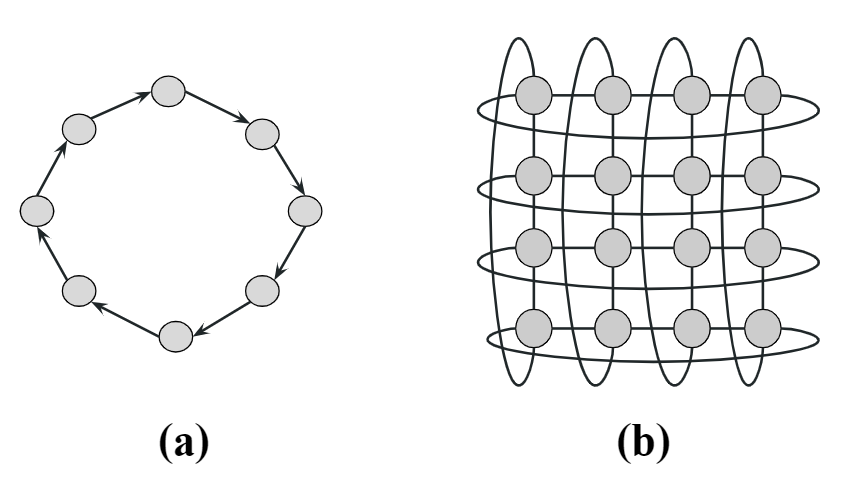}
   \caption{Graph topologies used in the decentralized experiments: (a) 8-node directed ring, and (b) 16-node (4x4) undirected Torus.}
   \label{fig:graphs}
\end{figure}

We evaluate the performance of the proposed low precision decentralized training on image recognition task using CIFAR-10 \cite{cifar10} and a subset of ImageNet \cite{imagenette} datasets. we train on ResNet \cite{resnet} and VGG architectures \cite{vgg} over a decentralized graph topologies of ring and Torus as shown in Figure.~\ref{fig:graphs}. Note that we modify the linear layers of VGG-11 to contain a hidden size of 512 and output size of 10.
We use stochastic gradient descent optimizer with Nesterov momentum and the momentum hyper-parameter is set to $\beta=0.9$. The input samples were normalized before training for all the experiments. For the vision datasets, we used the following transformations as data augmentation techniques using torchvision's transforms package: 1. \href{https://pytorch.org/vision/stable/generated/torchvision.transforms.RandomCrop.html}{Random crop} and 2. \href{https://pytorch.org/vision/stable/generated/torchvision.transforms.RandomHorizontalFlip.html} {Random horizontal flip}.

\begin{table}[t]
\centering
\caption{ \centering Baseline accuracy of training on a single node. Acc. (FP) indicates test accuracy (\%) with full precision training and Acc. (LP) indicates test accuracy with low precision (8-bit) training. }
\label{tab:baseline}
\resizebox{\columnwidth}{!}{
\begin{tabular}{|l|l|c|c|c|}
 \hline
Dataset & Model & Params & Acc. & Acc. \\
 & & (M) & (FP) & (LP) \\
\hline
CIFAR10&ResNet-20&0.27&91.25&91.12\\
\hline
CIFAR10&ResNet-54&0.76&91.99&91.66\\
\hline
CIFAR10&VGG-11&9.49&89.59&89.78\\
\hline
Imagenette&ResNet-18& 11.2 &87.26 & 87.26\\
\hline
Imagenette&VGG-11&9.49&89.58&87.39\\
\hline
AGNews&DistillBERT & 66.7 & 94.47 & 93.89\\
\hline
Sentlen&DistillBERT & 66.7 & 99.17 & 99.36 \\
\hline
 \end{tabular}
 }
\end{table}

\begin{figure*}[t]
  \centering
   \includegraphics[width=\linewidth]{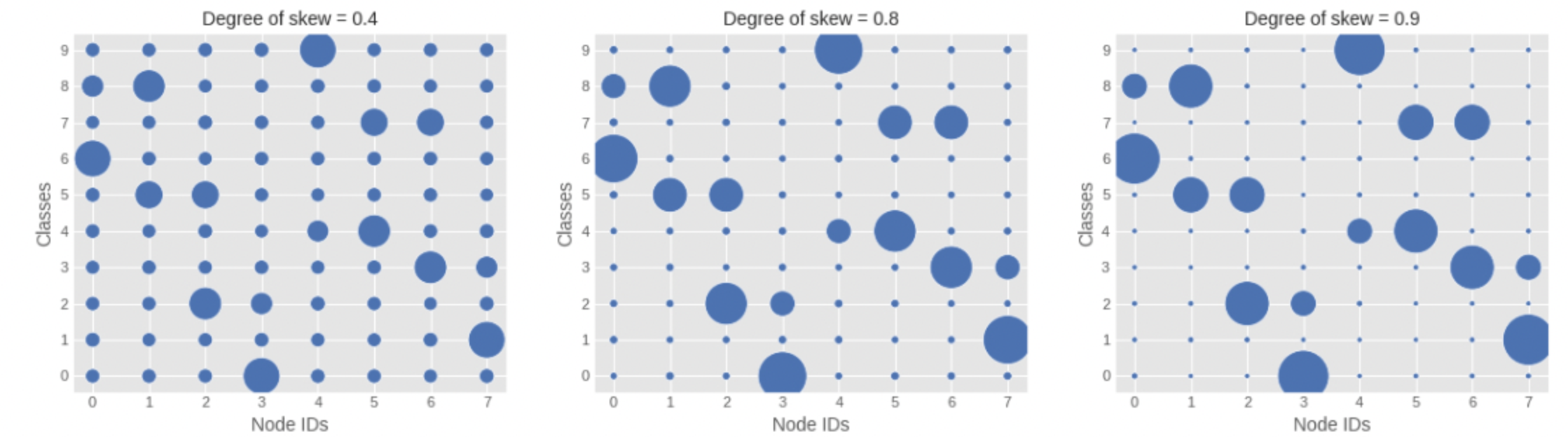}
   \caption{The distributions of CIFAR-10 classes across 8 nodes in a ring topology with various degree of skew in the partition.}
   \label{fig:skew}
   \vspace{1mm}
\end{figure*}

\begin{table*}[ht!]
\centering
\caption{ \centering Decentralized training of CIFAR-10 over various graph topolgies and models. Data column shows the data transferred by each node during training. Comp. indicates the percentage of sparsification. FP: Full Precision (32bit), LP: Low Precision (8bit).}
\label{tab:cifar}
\begin{tabular}{|p{2.4cm}|p{0.8cm}|p{1.0cm}|p{1.5cm}|p{1.5cm}|p{1.3cm}|p{1.3cm}|p{1.3cm}|p{1.5cm}|}
 \hline
Model/& Nodes  & Comp.&Data & Degree of &  DS & LP-DS & Choco & LP-Choco \\
Graph&(n)& ($\%$)&(GB) & skew & (32-bit) & (8-bit) & (32-bit) & (8-bit)\\
\hline
&&&&0&90.56&90.57&\textbf{90.96}&90.30\\
&&0&42.85 (FP)&0.4&90.36&89.89&\textbf{90.37}&90.29\\
&&&10.71 (LP)&0.8&87.57&87.68&\textbf{87.87}&87.06\\
&8&&&0.9&83.82&82.94&83.09&\textbf{83.65}\\
\cline{3-9}
&&&&0&88.67&87.58&\textbf{89.21}&88.23\\
&&99&0.651 (FP)&0.4&\textbf{88.64}&88.02&88.58&88.44\\
&&&0.163 (LP)&0.8&85.43&84.44&\textbf{85.78}&84.79\\
ResNet-20/&&&&0.9&82.94&81.92&\textbf{83.49}&82.22\\
\cline{2-9}

Directed Ring&&&&0&89.92&89.30&90.23&\textbf{90.26}\\
&&0& 21.21 (FP)&0.4&88.88&87.85&\textbf{90.01}&89.59\\
&&& 5.301 (LP)&0.8&85.21&85.39&\textbf{86.47}&86.47\\
&16&&&0.9&77.16&77.57&78.70&\textbf{79.53}\\
\cline{3-9}
&&&&0&86.58&85.45&\textbf{88.43}&87.72\\
&&90&3.177 (FP)&0.4&85.92&84.67&\textbf{87.98}&86.85\\
&&&0.794 (LP)&0.8&80.60&79.44&\textbf{82.76}&80.24\\
&&&&0.9&76.44&73.69&\textbf{77.42}&74.66\\
\hline

&&&&0&91.33&\textbf{92.09}&91.80&91.44\\
&&0&119.2 (FP)&0.4&91.40&91.28&\textbf{91.45}&91.19\\
&&&29.81 (LP)&0.8&89.72&88.98&\textbf{90.19}&89.30\\
ResNet-54/&8&&&0.9&87.15&84.06&87.17&\textbf{87.40}\\
\cline{3-9}
Directed Ring&&&&0&\textbf{90.23}&88.57&89.44&88.85\\
&&99&1.808 (FP)&0.4&89.48&88.60&\textbf{89.31}&88.81\\
&&&0.452 (LP)&0.8&85.91&84.69&\textbf{86.57}&85.06\\
&&&&0.9&\textbf{83.61}&81.93&83.27&81.87\\
 \hline
 
 &&&&0&89.64&89.13&\textbf{90.21}&89.73\\
&&0&744.5 (FP)&0.4&89.47&89.21&89.49&\textbf{89.80}\\
&&&186.1 (LP)&0.8&87.20&86.67&\textbf{87.21}&86.73\\
VGG-11/&16&&&0.9&86.26&85.15&\textbf{86.66}&85.67\\
\cline{3-9}
Undirected Torus&(4x4)&&&0&86.10&82.70&\textbf{86.72}&84.44\\
&&99&8.120 (FP)&0.4&85.42&81.53&\textbf{86.01}&83.49\\
&&&2.030 (LP)&0.8&80.12&77.98&\textbf{81.25}&80.53\\
&&&&0.9&77.10&72.89&\textbf{77.94}&76.11\\
 \hline
 \end{tabular}
\end{table*}

\textit{CIFAR-10 Dataset:} CIFAR-10 is a $10$-class image classification dataset with $50,000$ train samples and $10,000$ test samples each with a resolution of $32\times32$. The train and test samples are distributed equally across the $10$ classes. 
The experiments were conducted on directed ring topology (one peer per node) with 8, 16 nodes and undirected Torus (four peers per node) topology with 16 nodes. Note that, the spectral gap of 16 node directed ring is same as 32 node undirected ring. 
Spectral gap is a measure for the connectivity of the graph based on the graph size and number of neighbours per node \cite{spectral, chocosgd}.
We use ResNet-20, ResNet-54 and VGG-11 for evaluating CIFAR-10 dataset and all the experiments were run for 200 epochs. We use an initial learning rate of 0.1 decayed by a factor of 10 at epoch 100, 150 for ResNet architectures and decayed by factor of 2 every 30 epochs for VGG-11. We experiment with four different degree of skews ($0, 0.4, 0.8, 0.9$) and the distribution of classes across the nodes for these skews is shown in Figure.~\ref{fig:skew}. Note that skew of $0$ represents IID data partitions across the nodes. A $0.0$ degree of skew indicates the number of training samples across all the class are equally distributed for all the nodes. Increasing the degree of skew makes the distribution at each node to be skewed towards one or more classes. The results on CIFAR-10 dataset are shown in Table.~\ref{tab:cifar}.

\begin{table}[ht!]
\centering
\caption{ \centering Results of training CIFAR-10 dataset on ResNet-20 architecture over 8-node directed topology using Quasi-Global Momentum and Choco-SGD (both low and full precision).}
\label{tab:qgm}
\begin{tabular}{|c|c|c|c|c|}
 \hline
Compression & Degree of & Choco & LP-Choco \\
 (\%) & skew & with QGM & with QGM \\
\hline
&0.0&91.07&90.20\\
0&0.4&90.64&90.10\\
&0.8&87.64&87.16\\
&0.9&84.49&\textbf{85.28}\\
\hline
&0.0&89.87&89.12\\
99&0.4&89.39&88.73\\
&0.8&86.76&85.52\\
&0.9&84.42&83.68\\
\hline
 \end{tabular}
\end{table}

\begin{figure}
\centering     
\includegraphics[width=80mm]{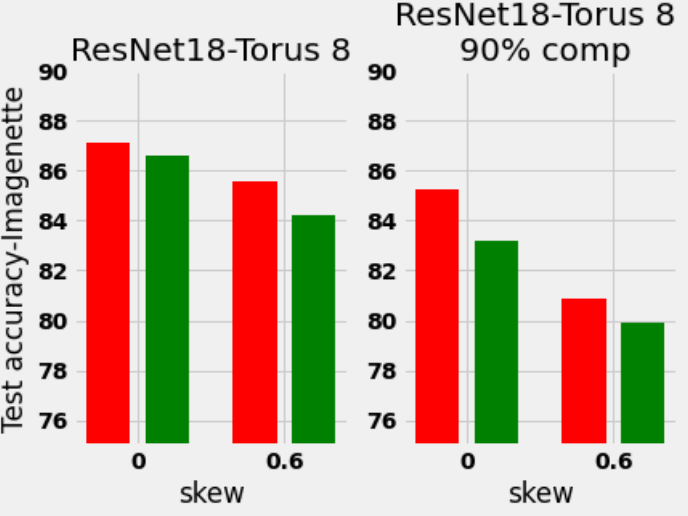}
\caption{Test accuracy with different skew for ResNet18 model trained on Imagnette. Red bars are models trained with Full precision and Green bars indicate the models trained with low precision.}
\label{fig:choco-sgd_imagenette}
\end{figure}

\textit{Imagenette Dataset:} The scalability of the proposed low precision decentralized training is shown by evaluating on imagenette.
Imagenette is a subset of Imagenet and is a $10$-class image classification dataset with $9,469$ training samples and $3,925$ test samples each, with a resolution of $224 \times 224$.
The train and test samples are distributed nearly equally across the $10$ classes.
Even though Imagenette has 10 classes, it is a complex dataset compared to CIFAR-10 because of the higher resolution and lower training set size.
The experiments were conducted on directed ring topology (one peer per node) with 4 nodes and on undirected Torus (four peers per node) topology with 8 nodes. We use ResNet-18 and VGG-11 architectures for evaluating this dataset and all the experiments were run for 100 epochs. An initial learning rate of 0.1 decayed by a factor of 10 at epoch 50, 75 was used for ResNet-18 and an initial learning rate of 0.01 decay by factor of 2 every 30 epochs for VGG-11. 
We experiment with two different degrees of skew --  0 (IID), 0.6 (non-IID). Since Imagenette has less samples per class in the training set (around 1000 per class), we used lower degree of skew (0.6) compared to CIFAR-10 to evaluate the Non-IID case. Note that a skew of 0.6 for imagenette has same impact as skew of 0.8 for CIFAR-10 in terms of data distribution. The results on imagenette dataset are shown in Table.~\ref{tab:imagenette} and Figure.~\ref{fig:choco-sgd_imagenette}

\begin{table}[ht!]
\centering
\caption{ \centering Decentralized training on Imagenette dataset (Subset of Imagenet with 10 classes) over different graph topologies.}
\label{tab:imagenette}
\begin{tabular}{|c|c|c|c|c|}
 \hline
Model/  & Comp. & Skew & Choco & LP-\\
Graph (n)& ($\%$) &  &  & Choco\\
& &  & (32-bit) & (8-bit)\\
\hline
&&0.0&87.10& 86.60\\
ResNet-18/&0&0.6&85.58&84.20\\
\cline{2-5}
Torus (8)&&0.0&85.25&83.17\\
&90&0.6&80.84&79.89\\
\hline
&&0.0&89.76&88.49\\
VGG-11/&0&0.6&88.89&86.82\\
\cline{2-5}
Ring (4) &&0.0&88.92&87.25\\
&90&0.6&88.20&86.27\\
\hline
 \end{tabular}
\end{table}

\begin{figure*}
\centering     
\subfigure[Choco-SGD without compression]{\label{fig:choco-sgd_wo_comp}\includegraphics[width=77mm]{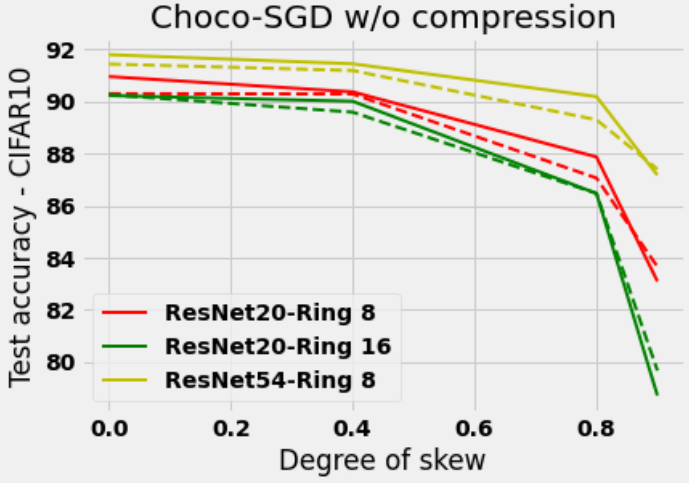}}
\subfigure[Choco-SGD without compression]{\label{fig:choco-sgd_comp}\includegraphics[width=80mm]{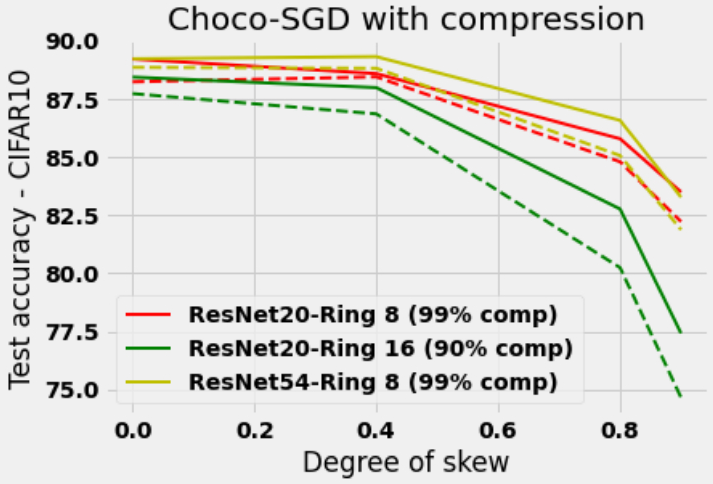}}
\caption{Variation of test accuracy with skew for different models and network configurations on CIFAR10 dataset. The solid lines are the plots for Full precision models and dashed lines are the plots for Low Precision models.}
\end{figure*}

Our experiments show that low precision decentralized training with the proposed Range EvoNorm has comparable accuracies in case of both IID and Non-IID data partitions. 
The low precision training with full communication (0\% compression) has minimal accuracy loss of less than $1\%$ compared to full precision training as shown in Figure.~\ref{fig:choco-sgd_wo_comp}. 
In the presence of communication compression, low precision training has $1-2\%$ loss in accuracy depending on the spectral gap and percentage of compression as shown in Figure.~\ref{fig:choco-sgd_comp}. 
From Table.~\ref{tab:cifar}, we note that CHOCO-SGD outperforms Deep-squeeze in almost all the cases.

\begin{figure}
\centering     
\includegraphics[width=80mm]{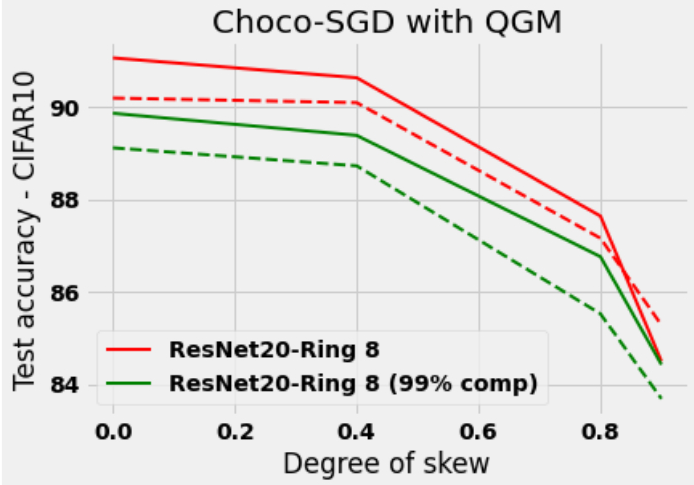}
\caption{Variation in test accuracy with skew for models trained with QGM on CIFAR10 dataset. The solid lines are the plots for Full precision models and dashed lines are the plots for Low Precision models.}
\label{fig:choco-sgd_qgm}
\end{figure}

We observe that for higher degree of skew (0.9), low precision training can performs better than full precision training with no communication compression. For CIFAR-10 dataset trained on ResNet-20, LP-Choco shows an improvement of $0.56\%$, $0.83\%$ over 8-node and 16-node directed ring topology, respectively. We hypothesise that this behaviour is due to the regularization effect of the low precision training. With higher skew values, the local models can easily overfit the local data and hence, regularizing the weights while training can result in better performance. We evaluate the proposed low precision decentralized training with Quasi-Global momentum \cite{qgm} shown in Table.~\ref{tab:qgm} and Figure.~\ref{fig:choco-sgd_qgm}. We observe that low precision training has similar improvements with skew as compared to full precision and hence, QG momentum can be used in synergy with our proposed training methodology.

In summary, we observed the following trends with low precision decentralized training:
1) The proposed low precision decentralized training trades-off of $1-2\%$ accuracy for $\sim 20\times$ reduction in compute and memory cost during training (Section \ref{sec:hardware})
2) Lower spectral gap can support higher communication compression with less than $1\%$ accuracy loss for low precision training over IID data for same number of epochs. 
Spectral gap is a measure for the connectivity of the graph based on the graph size and number of neighbours per node.
3) Communication compression results in higher degradation of performance for low precision decentralized learning over non-IID data compared to the IID counterpart.
4) Larger and redundant model architectures (such as VGG-11) have higher degradation in performance with communication compression compared to compact and optimal models with skip connections (ResNet-20).
5) For higher degree of skew (0.9), low precision training can perform slightly better than full precision training with no communication compression.

\subsection{Natural Language Processing Tasks}

\begin{table*}[t]
\centering
\caption{ \centering Decentralized training of various Natural Language Processing (NLP) tasks over a  directed 8 node ring topology.}
\label{tab:nlp}
\begin{tabular}{|p{2.5cm}|p{1.7cm}|p{1.5cm}|p{2.0cm}|p{2.0cm}|p{2.0cm}|p{2.0cm}|}
 \hline
Dataset/  & Compression & Degree of & DS & LP-DS & Choco & LP-Choco \\
Model& ($\%$) & skew & (32-bit) & (8-bit) & (32-bit) & (8-bit)\\
\hline
&&0.0&94.18&93.99&\textbf{94.21}&94.12\\
&0&0.5&94.13&93.72&\textbf{94.15}&93.61\\
AGNews/&&0.9&\textbf{92.19}&91.35&91.56&91.49\\
\cline{2-7}
DistillBERT&&0.0&93.88&93.55&\textbf{94.54}&94.04\\
&99&0.5&93.41&92.92&\textbf{93.83}&93.22\\
&&0.9&91.93&91.21&\textbf{92.33}&91.80\\
\hline
&&0.0&99.39& 98.83 &\textbf{99.48}&98.75\\
&0&0.5& 99.54 & 98.92 &\textbf{99.55}& 99.00\\
Sentlen/&&0.9& \textbf{99.39} & 98.46 & 99.33 & 98.32 \\
\cline{2-7}
DistillBERT&&0.0& 98.67 & 98.24 & \textbf{99.10} & 98.59 \\
&99&0.5 & 98.66 & 97.94 & \textbf{98.95} & 98.55\\
&& 0.9 & 98.04 & 97.28 & \textbf{98.72} & 97.73 \\
\hline
 \end{tabular}
\end{table*}

We analyze the effect of low precision training on natural language processing task for AG News \cite{Zhang2015CharacterlevelCN} and Sentlen from the probing dataset in SENEVAL framework \cite{conneau2018senteval} with DistillBERT \cite{sanh2019distilbert} as our base model. Our experimental setup consists of 4 nodes connected through directed ring topology. We evaluate low precision training with Deep-Squeeze and Choco-SGD for three different degrees of skews ($0.0, 0.5, 0.9$) and two different compression ratios ($0\%$, $99\%$ ). Both the datasets are fine tuned for 5 epochs with Adam optimizer and a learning rate, $\beta_1$, $\beta_2$ and weight decay of $3 \times 10 ^{-5}$, $0.9$, $0.9$, and $1 \times 10 ^{-4}$ respectively.

\textit{AG News Dataset:} AG News dataset \cite{Zhang2015CharacterlevelCN} is a 4 way classification dataset constructed by assembling titles and description fields of the articles from the 4 largest classes (`Sports',`Sci/Tech',`Business',`World') of AG's News Corpus \footnote{\url{http://groups.di.unipi.it/~gulli/AG_corpus_of_news_articles.html}}. The dataset consists of 120000 training samples and 7600 test samples distributed equally across the 4 classes. 

\textit{Sentlen task:} Sentlen task from the probing dataset in the SentEVAL framework \cite{conneau2018senteval} is a 6 way classification task with the goal to predict the length of the sentences. The 6 classification bins are formed by sentence lengths belonging to the intervals 5-8, 9-12, 13-16, 17-20 21-25 and 26-28. The task consists of 100,000 training and 10,000 validation samples distributed nearly equally across the 6 classes.

The results for the NLP experiments are presented in Table.~\ref{tab:nlp}. Low precision training has less than $1\%$ drop in accuracy as compared to full precision counterpart for both Deep-Squeeze and Choco-SGD. We do not observe the regularization effect of low precision training for higher degree of skew ($0.9$) as seen in Vision tasks. This can be potentially due to the large size of the model masking the effects of regularization through low precision training. 

\section{Hardware Implications}\label{sec:hardware}
\begin{table*}[t]
\centering
\caption{ \centering  Number of multiplication and additions involved in different layers of the network during training (Forward+backward pass) in terms of input activation maps size, number of output channels, kernel size, padding stride and input resolution.}
\label{tab:hardware_training}
\resizebox{\textwidth}{!}{
\begin{tabular}{cc}
     \begin{minipage}{\linewidth}
     \centering
    \resizebox{\textwidth}{!}{
     \begin{tabular}{|c|c|c|}
         \hline
        Layer  & Number Of & Number Of \\
        type   &Additions & Multiplications \\
        \hline
        \multirow{2}*{Conv}& \multirow{2}*{$3 \times b \times C_{out} \times C_{in} \times K^{2} \times \left[\frac{L+2 \times P - K}{S} + 1\right]^{2}$} & \multirow{2}*{$3 \times b \times C_{out} \times C_{in} \times K^{2} \times \left[\frac{L+2 \times P - K}{S} + 1\right]^{2}$}\\
        ~&~ &~\\
        \hline
        Linear&  $3 \times b \times C_{out} \times C_{in}$ &$3 \times b \times C_{out} \times C_{in}$ \\
        \hline
        Quantization(Activation)&$2 \times b \times C_{in} \times L^{2}$&$4 \times b \times C_{in} \times L^{2}$ \\
        \hline
        Quantization(Weight)& $2 \times C_{out} \times C_{in} \times K^{2}$&$4 \times C_{out} \times C_{in} \times K^{2}$ \\
        \hline
        Range batch-&\multirow{2}*{$9 \times b \times C_{in} \times L^{2}$} &\multirow{2}*{$6 \times b \times C_{in} \times L^{2} + 7 \times C_{in}$ }\\
        norm+ReLU & ~&~ \\
        \hline
        Range EvoNorm& $6 \times b \times C_{in} \times L^{2}$ &$10 \times b \times C_{in} \times L^{2} + 5 \times C_{in}$ \\
         \hline
        Skip Connections& $2 \times b \times C_{in} \times L^{2}$ &-\\
        \hline
        Average Pool& $ b \times C_{in} \times L^{2}$ & $2 \times b \times C_{in}$\\
        \hline
         \end{tabular}
         }
    \end{minipage} 
    
 \end{tabular}
 }
\end{table*}

\begin{table}[]
    \centering
        \caption{Description for the notations used in the equations in Table \ref{tab:hardware_training} and \ref{tab:hardware_inf}} 
    \label{tab:description}
         
    \begin{minipage}{\linewidth}
     \centering
    \resizebox{\textwidth}{!}{
     \begin{tabular}{|c|c|}
         \hline
        Notation  & Description \\
        \hline
        $b$  & Batch size \\
        \hline
        $C_{in}$  &\# of Input channels \\
        \hline
         $C_{out}$  &\# of Output channels \\
        \hline
        $K$  & Kernel Size for convolution \\
        \hline
        $L$  & Height/Width of input \\
        \hline
        $P$  & Padding for convolution \\
        \hline
        $S$  & Stride for convolution \\
        \hline
         \end{tabular}
        }
    \end{minipage} 
\end{table}

\begin{figure}[t]
  \centering
   \includegraphics[width=\linewidth]{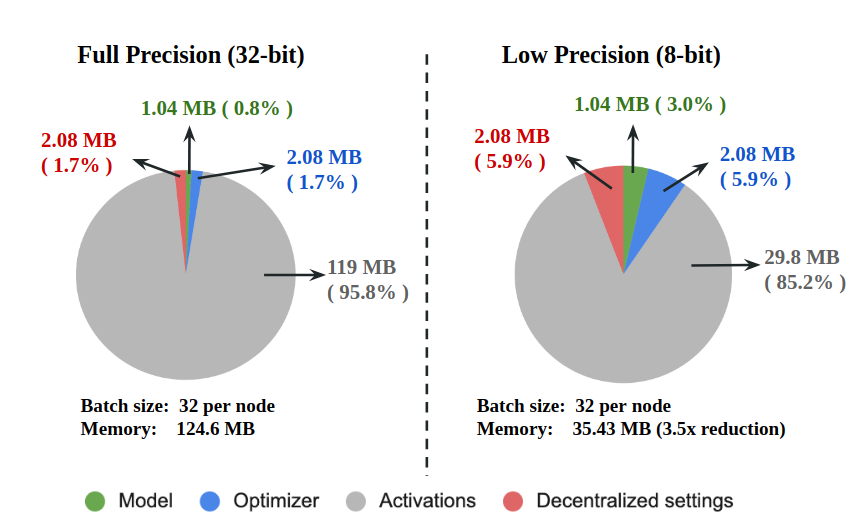}
   \caption{Memory and compute requirements for each node per iteration of full precision vs low precision training of CIFAR-10 dataset on ResNet-20 architecture over a decentralized setup.}
   \label{fig:hardware}
\end{figure}

Let us discuss the energy and memory benefits of low precision training. 
All our experiments were conducted on a system with Nvidia GTX 1080ti GPU.
During the training of a deep learning model, majority of the memory usage comes from storing the activations of the batch of images to compute the gradients. Data-parallel distributed learning reduces this memory requirement by lowering the batch-size per device, which in turn decreases the computations per iteration while still keeping the throughput constant.
However, this has no effect on network size and each node has to still store and access the millions of parameters of a deep neural network.
The network size can be reduced by employing low precision decentralized training which quantizes the weights and activations to 8-bits. This further reduces the total compute and memory requirement enabling training on resource constrained edge devices such as drones, smartphones etc. Figure.~\ref{fig:hardware} show the distribution of memory during full precision and low precision training. Note that, weights are stored in full precision during training and in low precision for inference.  

We evaluate the energy benefits of the network by computing the total number of add and multiply operations required in the training and inference. 
In particular, we use integer quantization which results in integer operations for our low precision models and float operations for the baseline models. 
The energy for 32 bit floating point add, 32 bit floating point multiply, 8 bit integer add and 8 bit integer multiply were taken from \cite{horowitz20141}. Table \ref{tab:hardware_training} and Table \ref{tab:hardware_inf} show the break down of number of add and multiply operations required by different layers in terms of the batch size $b$,input channels $C_{in}$, output channels $C_{out}$, input resolution $L$, kernel size $K$, stride $S$ and padding $P$. These parameters are also described in Table \ref{tab:description}. Our analysis shows we roughly get $20\times$ energy benefits for our low precision implementations as compared to the full precision, shown in Table \ref{tab:hardware_results}. 

We also compute the memory requirement for both full precision and low precision training (reported in Table.~\ref{tab:hardware_results}). An example for distribution of memory during training for a decentralized setup is shown in Figure.~\ref{fig:hardware}. For inference, we report the memory utilized by the model weights. For model architectures such as ResNet where majority of memory usage during training comes from activations, we observe a $\sim 3.5\times$ reduction in memory usage with low precision training. But for parameter heavy networks such as VGG, the benefits in memory are reduced to $1.3\times$. Even though `Range batch-norm' shows more efficiency in terms of memory than `Range EvoNorm' with low precision training, the absolute memory utilization is less with `Range EvoNorm'.
In summary our approach reduces the communication cost by $4\times$, compute cost by $\sim 20 \times$ and memory requirements by $\sim 3.5 \times$.


\begin{table*}[t]
\centering
\caption{ \centering (left) Number of multiplication and additions involved in different layers of the network during inference (Forward pass) in terms of input activation maps size, number of output channels, kernel size, padding stride and input resolution. }
\label{tab:hardware_inf}
\resizebox{\textwidth}{!}{
\begin{tabular}{cc}
     \begin{minipage}{\linewidth}
     \centering
    \resizebox{\textwidth}{!}{
     \begin{tabular}{|c|c|c|}
         \hline
        Layer  & Number Of & Number Of \\
        type   &Additions & Multiplications \\
        \hline
        \multirow{2}*{Conv}& \multirow{2}*{$b \times C_{out} \times C_{in} \times K^{2} \times \left[\frac{L+2 \times P - K}{S} + 1\right]^{2}$} & \multirow{2}*{$b \times C_{out} \times C_{in} \times K^{2} \times \left[\frac{L+2 \times P - K}{S} + 1\right]^{2}$}\\
        ~&~ &~\\
        \hline
        Linear&  $b \times C_{out} \times C_{in}$ &$b \times C_{out} \times C_{in}$ \\
        \hline
         Quantization(Weight)&$2 \times C_{out} \times C_{in} \times K^{2}$&$4 \times C_{out} \times C_{in} \times K^{2}$ \\
        \hline
        Range batch-&\multirow{2}*{$3 \times b \times C_{in} \times L^{2}$} &\multirow{2}*{$2 \times b \times C_{in} \times L^{2} + 3 \times C_{in}$ }\\
        norm+ReLU & ~&~ \\
        \hline
         Range EvoNorm& $2 \times b \times C_{in} \times L^{2}$ &$4 \times b \times C_{in} \times L^{2} + 2 \times C_{in}$ \\
        \hline
         Skip Connections& $2 \times b \times C_{in} \times L^{2}$ &-\\
        \hline
        Average Pool& $ b \times C_{in} \times L^{2}$ & $ b \times C_{in}$\\
        \hline
         \end{tabular}
         }
    \end{minipage} 
 \end{tabular}
 }
\end{table*}

\begin{table*}[ht!]
\centering
\caption{ \centering Energy efficiency of low precision (8bit) compared to full precision (32bit) for each node (batch-size: 32) per iteration during training and inference over CIFAR-10. RBN: Range batch-norm, REN: Range EvoNorm.}
\label{tab:hardware_results}
\begin{tabular}{|c|c|c|c|c|c|c|c|}
 \hline
Model  & Norm & \multicolumn{3}{c|}{Memory (MB)} & \multicolumn{3}{c|}{Energy (mJ)} \\
\cline{3-8}
& type& Choco & LP-Choco & Efficiency & Choco & LP-Choco & Efficiency\\
\cline{3-8}
& & \multicolumn{6}{c|}{Training}\\
\hline
\multirow{2}*{ResNet-20}&RBN & 170.60& 46.15& 3.70& 68.75 &3.45 &19.93\\
 ~& REN &124.60 &35.43 & 3.52&68.82 & 3.45&19.93\\
\hline
\multirow{2}*{ResNet-54} &RBN& 424.85& 117.05& 3.63& 392.55& 19.69&19.94\\
~& REN & 308.85& 88.05& 3.51& 392.95& 19.71&19.9\\
\hline
\multirow{2}*{VGG-11}&RBN &300.48 &210.95 & 1.42& 173.56&8.69 &19.97\\
 & REN &263.48 & 201.70& 1.31&173.61 &8.70 &19.67\\

\cline{3-8}
& & \multicolumn{6}{c|}{Inference}\\
\hline
\multirow{2}*{ResNet-20}&RBN &1.04 &0.26 &4 &22.92 & 1.16&19.93\\
 ~& REN & 1.04&0.26 &4 &22.96 & 1.16&19.78\\
\hline
\multirow{2}*{ResNet-54} &RBN&2.89 & 0.72&4 &130.85 &6.60 & 19.82\\
~& REN &2.89 & 0.72&4 & 131.07 & 6.61 &19.82\\
\hline
\multirow{2}*{VGG-11}&RBN & 36.22&9.01 &4 & 57.85& 2.90&19.96\\
 & REN &36.22 &9.01 &4 &57.88 & 2.90&19.96\\
\hline
 \end{tabular}
\end{table*}

\section{Limitations}
From the series of experiments with various datasets, architectures and graph topologies, we found the following limitations with communication and compute efficient decentralized learning. The performance (accuracy) decreases with increase in the degree of skew in the data partitions across the nodes. When low precision training is accompanied by communication compression, it incurs $\sim 1-2\%$ more degradation in accuracy (in most cases) compared to full communication. Apart from data and model complexity, various decentralized graph parameters such as spectral gap, degree of skew and percentage of communication compression also play a vital role in determining the performance of decentralized learning task. 
In most of the cases, the models do not converge (for both full precision and low precision training) for a skew of $1$ ($100\%$ non-IID) i.e., every node has data from a unique set of classes.
Choco-SGD performs better than deep-squeeze but needs additional memory buffer on each node with size equivalent to model size.

\section{Conclusion}
Efficient decentralized distributed learning with practical constraints is the key to training intelligent deep learning models on edge devices. In this paper, we propose low precision decentralized training and conduct a detailed study to reveal and understand the effect of various factors on the training. 
We demonstrate the convergence of decentralized low precision training under the variation of several key factors such as the learning algorithm (Choco-SGD and Deep-Squeeze), degree of skew and communication compression. We evaluated the effect of all the above factors over a considerable number of tasks, datasets, models and graph topologies. The following are the key findings from our study:
1) Our experiments show that low precision decentralized learning converges with minimal loss in accuracy compared to full precision training. This reduces the communication during training by $4\times$ and compute during training and inference by $\sim 20 \times$. 
2) A suitable normalization activation layer that works better with both quantization and non-IID data is `Range EvoNorm'. 
3) Decentralized algorithms that communicate the gradients (such as Choco-SGD) at the cost of additional memory buffer, perform better in terms of accuracy than algorithms that communicate the weights such as Deep-Squeeze. 
4) We observe that there is a slight improvement in accuracy for low precision training compared to full precision in the presence of high skew ($0.9$) and no communication compression. 
5) We also show the effect of communication compression with increasing skew for both full precision and low precision training. 
We hope that the findings and insights in this paper will help in better understanding of efficient decentralized learning over non-IID data and spur further research in the fundamental problem of non-IID data in decentralized learning.

\section{Future Work}
We plan to focus on making algorithmic changes to improve the performance of the decentralized learning over non-IID data distribution. There is a huge gap in performance of decentralized learning between IID and non-IID distribution \cite{skewscout, qgm, cga} and it gets worse with increase in graph size of the decentralized setup. So, we plan to explore gradient manipulation techniques similar to \cite{cga} but with minimal communication overheads. \cite{cga} propose Cross Gradient Aggregation (CGA) technique to improve performance with non-IID data but at a cost of increasing the communication by $2\times$. The authors propose compressed version of the algorithm (CompCGA) trading off on the performance. Also, this paper only presents the results on CIFAR-10 dataset with a 5-layered CNN and VGG architectures. 
We will be looking at alternative gradient manipulation techniques using feature importance methods such as Principal Component Analysis (PCA) that could help us improve the performance of non-IID decentralized learning without resulting in immense communication overheads. We also plan to analyse the feasibility of such gradient manipulation techniques on deep networks (ResNets, Xception etc.) and complex datasets such as ImageNet.

\section*{Acknowledgement}
This work was supported in part by the Center for Brain Inspired Computing (C-BRIC), one of the six centers in JUMP, a Semiconductor Research Corporation (SRC) program sponsored by DARPA, by the Semiconductor Research Corporation, DARPA ShELL project, the National Science Foundation, Intel Corporation, the DoD Vannevar Bush Fellowship, and by the U.S. Army Research Laboratory and the U.K. Ministry of Defence under Agreement Number W911NF-16-3-0001.

\newpage
\bibliographystyle{cas-model2-names}

\bibliography{cas-refs}

\bio{}
\endbio

\bio{}
\endbio

\end{document}


\def\floatpagepagefraction{1}
\def\textpagefraction{.001}



\title{Supplementary for Low Precision Decentralized Distributed Training with Heterogeneous Data}  
\date{\vspace{-5ex}}

\author{Sai Aparna Aketi, Sangamesh Kodge, Sangamesh Kodge}
\maketitle

\appendix
\section{Algorithmic Details}

Algorithm.~\ref{alg:deepsqueeze} describes the Deep-Squeeze algorithm proposed in \cite{deepsqueeze} and Algorithm.~\ref{alg:chocosgd} describes CHOCO-SGD proposed in \cite{choco2}. Let $G(V,E)$ be the decentralized graph topology and $W$ is the mixing matrix (adjacency matrix) of of the graph. The graph $G$ contains $n$ nodes i.e., $|V| = n$ and the size of $W$ is $n\times n$.
\setcounter{algorithm}{1}
\begin{algorithm}[ht!]
\textbf{Input:} On each node $i\in [n]$ - initialize model parameters $x_0^{(i)}$, initialize error $\delta_0^{(i)}$ to $0$s, learning rate $\gamma$, averaging rate $\eta$, mixing matrix $W$, and compression operator $C$.\\
1. \textbf{for} $t=0,\hdots,T-1$, at each node $i$, \textbf{do} \\
   \hspace*{6mm}Each node simultaneously implements:\\
2.\hspace{3mm}Randomly sample a mini-batch $\xi_t^{(i)}$ from local\\ 
   \hspace*{6mm} distribution $D_{i}$\\
3.\hspace{3mm}Compute local gradient $g_t^{(i)} = \nabla F_i(x_t^{(i)}, \xi_t^{(i)}) $\\
4.\hspace{3mm}Do the local update $\Tilde{x}_{t}^{(i)} = x_t^{(i)}-\gamma g_t^{(i)}$\\
5.\hspace{3mm}Compute error compensated update $v_{t}^{(i)}=\Tilde{x}_{t}^{(i)}+\delta_{t}^{(i)}$\\
6.\hspace{3mm}Compress the error compensated variable $C[v_{t}^{(i)}]$\\
7.\hspace{3mm}Update error $\delta_{t}^{(i)} = v_t^{(i)} - C[v_{t}^{(i)}]$\\ 
8.\hspace{3mm}\textbf{for} each neighbour $j$ of $i^{th}$ node ($j \in \mathcal{N}_{i}$) \textbf{do}\\
   \hspace*{9mm} Send $C[v_{t}^{(i)}]$ and receive $C[v_{t}^{(j)}]$  \\
9.\hspace{3mm}\textbf{end for}\\
10.\hspace{3mm}Do the gossip update for local models:\\
   \hspace*{6mm} $x_{t+1}^{(i)} = \Tilde{x}_{t}^{(i)}+ \eta \sum_{j\in \mathcal{N}_i}(W_{ij}-I_{ij})C[v_{t}^{(j)}]$\\
11. \textbf{end for}\\
\caption{Deep-Squeeze \cite{deepsqueeze}}
\label{alg:deepsqueeze}
\end{algorithm}

\begin{algorithm}[ht!]
\textbf{Input:} On each node $i\in [n]$ - initialize model parameters $x_0^{(i)}$, initialize $\hat{x}_0^{(j)} \hspace{1mm}\forall j \in N(i)$  to $0$s, learning rate $\gamma$, averaging rate $\eta$, mixing matrix $W$, and compression operator $C$.\\

1. \textbf{for} $t=0,\hdots,T-1$, at each node $i$, \textbf{do} \\
   \hspace*{6mm}Each node simultaneously implements:\\
2.\hspace{3mm}Randomly sample a mini-batch $\xi_t^{(i)}$ from local\\ 
   \hspace*{6mm} distribution $D_{i}$\\
3.\hspace{3mm}Compute local gradient $g_t^{(i)} = \nabla F_i(x_t^{(i)}, \xi_t^{(i)}) $\\
4.\hspace{3mm}Do the local update $\Tilde{x}_{t}^{(i)} = x_t^{(i)}-\gamma g_t^{(i)}$\\
5.\hspace{3mm}Compute and compress the sending information: 
\hspace*{6mm} $q_{t}^{(i)}=C[\Tilde{x}_{t}^{(i)}-\hat{x}_{t}^{(i)}$]\\
7.\hspace{3mm}\textbf{for} each neighbour $j$ of $i^{th}$ node ($j \in \mathcal{N}_{i}$) \textbf{do}\\
   \hspace*{9mm} Send $q_{t}^{(i)}$ and receive $q_{t}^{(j)}$  \\
   \hspace*{9mm} $\hat{x}_{t+1}^{(j)}$= $q_{t}^{(j)}+\hat{x}_{t}^{(j)} $  \\
8.\hspace{3mm}\textbf{end for}\\
9.\hspace{3mm}Do the gossip update for local models:\\
   \hspace*{6mm} $x_{t+1}^{(i)} = \Tilde{x}_{t}^{(i)}+ \eta \sum_{j\in \mathcal{N}_i}(W_{ij}-I_{ij})(\hat{x}_{t+1}^{(j)}-\hat{x}_{t+1}^{(i)})$\\
11. \textbf{end for}\\
\caption{CHOCO-SGD \cite{choco2}}
\label{alg:chocosgd}
\end{algorithm}

\begin{algorithm}[ht!]
\textbf{Input:} On each node $i\in [n]$ - initialize model parameters $x_0^{(i)}$, $z_0^{(i)}$ = $x_0^{(i)}$, initialize error $\delta_0^{(i)}$ to $0$s and bias weight $u_0^{(i)}$ to $1$, learning rate $\gamma$, averaging rate $\eta$, mixing matrix $W$, and compression operator $C$.\\

1. \textbf{for} $t=0,\hdots,T-1$, at each node $i$, \textbf{do} \\
   \hspace*{6mm}Each node simultaneously implements:\\
2.\hspace{3mm}Randomly sample mini-batch $\xi_t^{(i)}$ from local\\ 
   \hspace*{6mm} distribution $D_{i}$\\
3.\hspace{3mm}Compute local gradient $g_t^{(i)} = \nabla F_i(z_t^{(i)}, \xi_t^{(i)}) $\\
4.\hspace{3mm}Do the local update $\Tilde{x}_{t}^{(i)} = x_t^{(i)}-\gamma g_t^{(i)}$\\
5.\hspace{3mm}Compute error compensated update $v_{t}^{(i)}=\Tilde{x}_{t}^{(i)}+\delta_{t}^{(i)}$\\
6.\hspace{3mm}Compress the error compensated variable $C[v_{t}^{(i)}]$\\
   \hspace*{6mm} and update error $\delta_{t}^{(i)} = v_t^{(i)} - C[v_{t}^{(i)}]$\\ 
7.\hspace{3mm}Send $C[v_{t}^{(i)}]$ and $u_t^{(i)}$ to out-neighbours of $i^{th}$ node\\
\hspace*{6mm}(i.e. to $j \in \mathcal{N}_{i}^{out}$) \\
8.\hspace{3mm}Receive $C[v_{t}^{(j)}]$ and $u_t^{(j)}$ from all the in-neighbours\\  
   \hspace*{6mm} of $i^{th}$ node ($j \in \mathcal{N}_{i}^{in}$)\\
9.\hspace{3mm}Do the gossip update for local models and bias weights:
   \hspace*{6mm} $x_{t+1}^{(i)} = \Tilde{x}_{t}^{(i)}+ \eta \sum_{j\in \mathcal{N}_i^{in}}(W_{ij}-I_{ij})C[v_{t}^{(j)}]$\\
   \hspace*{6mm} $u_{t+1}^{(i)} = u_{t}^{(i)}+ \eta \sum_{j\in \mathcal{N}_i^{in}}(W_{ij}-I_{ij})u_{t}^{(j)}$\\
10.\hspace{3mm}De-bias the updated model: $z_{t+1}^{(i)} = \frac{x_{t+1}^{(i)}}{u_{t+1}^{(i)}}$\\
11. \textbf{end for}\\
\caption{Sparse-Push \cite{sparsepush} (Deep-Squeeze with SGP)}
\label{alg:sparsePush}
\end{algorithm}

\begin{algorithm}[ht!]
\textbf{Input:} On each node $i\in [n]$ - initialize model parameters $x_0^{(i)}$, $z_0^{(i)}$ = $x_0^{(i)}$, initialize $\hat{x}_0^{(j)} \hspace{1mm}\forall j \in N(i)$  to $0$s and bias weight $u_0^{(i)}$ to $1$, learning rate $\gamma$, averaging rate $\eta$, mixing matrix $W$, and compression operator $C$.\\

1. \textbf{for} $t=0,\hdots,T-1$, at each node $i$, \textbf{do} \\
   \hspace*{6mm}Each node simultaneously implements:\\
2.\hspace{3mm}Randomly sample mini-batch $\xi_t^{(i)}$ from local\\ 
   \hspace*{6mm} distribution $D_{i}$\\
3.\hspace{3mm}Compute local gradient $g_t^{(i)} = \nabla F_i(z_t^{(i)}, \xi_t^{(i)}) $\\
4.\hspace{3mm}Do the local update $\Tilde{x}_{t}^{(i)} = x_t^{(i)}-\gamma g_t^{(i)}$\\
5.\hspace{3mm}Compute and compress the sending information: 
\hspace*{6mm} $q_{t}^{(i)}=C[\Tilde{x}_{t}^{(i)}-\hat{x}_{t}^{(i)}$]\\
7.\hspace{3mm}Send $q_{t}^{(i)}$ and $u_t^{(i)}$ to out-neighbours of $i^{th}$ node ($\mathcal{N}_{i}^{out}$)
7.\hspace{3mm}\textbf{for} each neighbour $j$ of $i^{th}$ node ($j \in \mathcal{N}_{i}^{in}$) \textbf{do}\\
   \hspace*{9mm} Receive $q_{t}^{(j)}$ and $u_t^{(j)}$  \\
   \hspace*{9mm} $\hat{x}_{t+1}^{(j)}$= $q_{t}^{(j)}+\hat{x}_{t}^{(j)} $  \\
8.\hspace{3mm}\textbf{end for}\\
9.\hspace{3mm}Do the gossip update for local models and bias weights:
   \hspace*{6mm} $x_{t+1}^{(i)} = \Tilde{x}_{t}^{(i)}+ \eta \sum_{j\in \mathcal{N}_i^{in}}(W_{ij}-I_{ij})(\hat{x}_{t+1}^{(j)}-\hat{x}_{t+1}^{(i)})$\\
   \hspace*{6mm} $u_{t+1}^{(i)} = u_{t}^{(i)}+ \eta \sum_{j\in \mathcal{N}_i^{in}}(W_{ij}-I_{ij})u_{t}^{(j)}$\\
10.\hspace{3mm}De-bias the updated model: $z_{t+1}^{(i)} = \frac{x_{t+1}^{(i)}}{u_{t+1}^{(i)}}$\\
11. \textbf{end for}\\
\caption{Quant-SGP \cite{quant_sgp} (CHOCO-SGD with SGP)}
\label{alg:quantsgp}
\end{algorithm}

Both the above mentioned algorithms converge only for symmetric and doubly-stochastic mixing matrices.  This results in the following constraints: 1) $W_{ij}=W_{ji}$ ,  2) $\sum_{j} W{ij}=1 \hspace{1mm} \forall i$ and 3) $\sum_{i} W{ij}=1 \hspace{1mm} \forall j$ where $W_{ij}$ is the element in the $i^{th}$ row and $j^{th}$ column of $W$. The neighbours of each node $i$ is given by $N(i)$. Note that $N(i)$ includes $i$. $I_{ij}$ represents the element in the $i^{th}$ row and $j^{th}$ column of an identity matrix. The compressor operator $C$ is usually chosen as random/top-k quantization or sparification.
Deep-Squeeze and CHOCO-SGD use error feedback mechanism for compression forcing the compression operator to be unbiased. This allows both the algorithms to have very high compression rates such as $90\%, 99\%$. The averaging rate $\gamma$ is chosen based on the percentage of compression. $\gamma$ is $1$ for no compression and decreases with increase in compression. In the main paper we use pre-define functions $g,h$ to simply the explanation of decentralized training. For Deep-Squeeze, $v_{t}^{(i)}=g(\Tilde{x}_{t}^{(i)}) = \Tilde{x}_{t}^{(i)}+\delta_{t}^{(i)}$ and $h(C[v_{t}^{(j)}])=C[v_{t}^{(j)}]$. Similarly for CHOCO-SGD, we have  $v_{t}^{(i)}=g(\Tilde{x}_{t}^{(i)}) = \Tilde{x}_{t}^{(i)}-\hat{x}_{t}^{(i)}$ , $C[v_{t}^{(i)}] = q_{t}^{(i)}$ and $h(q_{t}^{(j)}) = \hat{x}_{t+1}^{(j)}-\hat{x}_{t+1}^{(i)}$.

In our experiments, we use the Sparse-Push version of Deep-Squeeze (shown in Algorithm.~\ref{alg:sparsePush}) and Quant-SGP version of CHOCO-SGD (shown in Algorithm.~\ref{alg:quantsgp}). Sparsh-Push and Quant-SGP requires the mixing matrix to be column stochastic ($\sum_{i} W{ij}=1 \hspace{1mm} \forall j$) removing the stronger constraint of symmetry and doubly stochasticity. This enables both the algorithms to converge for directed and time-varying graphs which is more practical.  


\newpage
\bibliographystyle{cas-model2-names}

\bibliography{cas-refs}